*Data and text mining*

# PhenoTagger: A Hybrid Method for Phenotype Concept Recognition using Human Phenotype Ontology

Ling Luo[1], Shankai Yan[1], Po-Ting Lai[1], Daniel Veltri[2], Andrew Oler[2], Sandhya Xirasagar[2], Rajarshi Ghosh[2], Morgan Similuk[2], Peter N. Robinson[3] and Zhiyong Lu[1,*]

[1]National Center for Biotechnology Information (NCBI), National Library of Medicine (NLM), National Institutes of Health (NIH), Bethesda, MD 20894, USA, [2]Bioinformatics and Computational Biosciences Branch, Office of Cyber Infrastructure and Computational Biology, National Institute of Allergy and Infectious Diseases, National Institutes of Health, Bethesda, MD 209892, USA, and [3]The Jackson Laboratory for Genomic Medicine, Farmington, CT 06032, USA.

*To whom correspondence should be addressed.

## Abstract
**Motivation:** Automatic phenotype concept recognition from unstructured text remains a challenging task in biomedical text mining research. Previous works that address the task typically use dictionary-based matching methods, which can achieve high precision but suffer from lower recall. Recently, machine learning-based methods have been proposed to identify biomedical concepts, which can recognize more unseen concept synonyms by automatic feature learning. However, most methods require large corpora of manually annotated data for model training, which is difficult to obtain due to the high cost of human annotation.
**Results:** In this paper, we propose PhenoTagger, a hybrid method that combines both dictionary and machine learning-based methods to recognize Human Phenotype Ontology (HPO) concepts in unstructured biomedical text. We first use all concepts and synonyms in HPO to construct a dictionary, which is then used to automatically build a distantly supervised training dataset for machine learning. Next, a cutting-edge deep learning model is trained to classify each candidate phrase (*n*-gram from input sentence) into a corresponding concept label. Finally, the dictionary and machine learning-based prediction results are combined for improved performance. Our method is validated with two HPO corpora, and the results show that PhenoTagger compares favorably to previous methods. In addition, to demonstrate the generalizability of our method, we retrained PhenoTagger using the disease ontology MEDIC for disease concept recognition to investigate the effect of training on different ontologies. Experimental results on the NCBI disease corpus show that PhenoTagger without requiring manually annotated training data achieves competitive performance as compared with state-of-the-art supervised methods.
**Availability:** The source code, API information and data for PhenoTagger are freely available at https://github.com/ncbi-nlp/PhenoTagger.
**Contact:** zhiyong.lu@nih.gov
**Supplementary information:** Supplementary data are available at *Bioinformatics* online.

## 1 Introduction

Phenotypes constitute the visible properties of an organism that are produced by the interaction of the genotype and the environment. A greater understanding of phenotype-disease associations can enhance disease diagnosis and treatment. Phenotype concept recognition aims to automatically extract the phenotype concept in biomedical ontologies from unstructured text, which is a fundamental step for further biomedical text mining. Human Phenotype Ontology (HPO) is an ontology that provides a standardized vocabulary of phenotypic abnormalities associated with 7,500+ diseases (Köhler *et al.*, 2019). It is used widely by researchers, clinicians, informaticians and electronic health record systems from around the world. Recently, the recognition of HPO concepts has received considerable attention. Despite a few attempts in the past, it remains a challenging task in biomedical text mining research due to the following reasons: Ambiguity, use of abbreviations, use of metaphorical expressions, use of hedging and various forms of qualifiers, complex intrinsic structure, and the fact that each component of a phenotype description may have a nested structure (Groza *et al.*, 2015).

In previous works, dictionary and machine learning-based methods have been attempted for HPO concept recognition and have exhibited promising results. Due to the lack of publicly available large manually annotated HPO concept corpora, most existing HPO concept recognition tools are based on dictionary methods. Examples of popular tools are the NCBO (National Center for Biomedical Ontology) annotator (Jonquet *et al.*, 2009), ZOOMA (Kapushesky *et al.*, 2012), the OBO (Open Biological and Biomedical Ontologies) annotator (Taboada *et al.*, 2014), SORTA (Pang *et al.*, 2015), Doc2Hpo (Liu *et al.*, 2019), and the Monarch Initiative platform (Shefchek *et al.*, 2020). These dictionary-based methods primarily construct the dictionary based on HPO and recognize the HPO concept using text matching technology. They can achieve high precision, as most of the concepts in the phenotype ontology are specific. Using exact entries in the dictionary, however, makes it difficult to effectively capture semantic and syntactic variants (i.e., synonyms) that are common in the literature but have not appeared in HPO. Therefore, dictionary-based methods typically suffer from low recall rates.

Recently, machine learning-based methods have exhibited great potential in automatic feature learning and have achieved state-of-the-art performance in several biomedical entity recognition tasks that involve genes/proteins (Wei *et al.*, 2015), diseases (Leaman *et al.*, 2013) and chemicals (Leaman *et al.*, 2015). Most involve the pipelined method, i.e., treating concept recognition as two separate tasks, named entity recognition and normalization (NER and NEN). First, NER aims to extract biomedical entity mentions from raw biomedical texts. Then, NEN is used to link mentions of the same entity together into a single concept of the biomedical ontology vocabulary. Unlike these well-studied tasks, research on machine learning in regard to HPO concept recognition is limited to rarely publicly available manually annotated corpora. For example, Lobo *et al.* (2017) present a system for identifying human phenotypes by combining the conditional random field (CRF) model and manual validation rules. This system, however, extracts only the phenotype entity mentions and does not link them to the HPO concept labels. Most machine learning methods require large corpora of annotated data for model training, which is difficult to obtain due to the high cost of human annotation, especially in the biomedical domain. A model trained on the small-scale training dataset often cannot generalize well to unseen concepts, but annotating a large-scale training dataset covering all classes of biomedical concepts is highly challenging (Baumgartner *et al.*, 2008).

More recently, Arbabi *et al.* (2019) proposed an ontology-guided neural concept recognizer (NCR) to identify clinical terms in medical text. NCR can efficiently identify unseen synonyms using a convolutional neural network (CNN)-based neural dictionary model and alleviate the problem of dependence on large-scale labeled training data. However, it does not take advantage of the dictionary-based method and does not consider overlapping concepts (i.e. the concepts that share at least one common token). The overlapping concepts are common in biomedical texts (for example, about 26% of the concepts are overlapping concepts in HPO Gold Standardized Corpus plus (GSC+) (Lobo *et al.*, 2017)), and they can be classified into two types. First, one concept is nested by the other. For example, a concept "severe mental retardation" (HP:0002187) contains a concept "mental retardation" (HP:0001249). Second, two concepts share some common tokens, but no one is completely contained by the other. For example, the concept "developmental abnormalities" (HP:0001263) and the concept "abnormalities of the eye" (HP:0000478) share the common token "abnormalities" in the text "developmental abnormalities of the eye".

To address these problems, we propose PhenoTagger, a hybrid phenotype recognition method that combines dictionary and machine learning-based methods to recognize HPO concepts in unstructured biomedical text. Different from previous methods, PhenoTagger contains the dictionary and deep learning-based tagger components. The dictionary-based tagger can rapidly and exactly match the terms in HPO. For the deep learning-based tagger, the BioBERT (Biomedical Bidirectional Encoder Representations from Transformers) (Lee *et al.*, 2020) model is trained using a distantly supervised training dataset built with HPO and the biomedical literature. It can efficiently identify unseen synonyms. Finally, the combined rules that take into account overlapping concepts are proposed to combine the results of the two taggers. In our experiments, PhenoTagger is evaluated on the HPO GSC+ and a newly constructed full-text corpus. The results show that PhenoTagger outperforms state-of-the-art methods. Further, our method can be applied easily to other biomedical concept recognition tasks with corresponding biomedical ontology. In particular, we retrained PhenoTagger for disease concept recognition using the disease ontology MEDIC (Davis *et al.*, 2012) and tested it on the NCBI disease corpus (Doğan *et al.*, 2014). The results show that PhenoTagger without requiring manually labeled training data achieves competitive performance as compared with state-of-the-art supervised methods, suggesting that our method is highly robust and generalizable for concept recognition solely based on a given ontology.

## 2 Methods

In this section, we described our hybrid method, which consists of three main components (i.e., dictionary-based tagger, deep learning-based tagger, and the combining of prediction results). The processing flowchart of PhenoTagger can be divided into two phases (i.e., training phase and test phase), as shown in Fig. 1. In the training phase, we first use all concept names, synonyms, and lemmas in HPO to construct a dictionary. Then, the dictionary and biomedical literature are employed to automatically build a distantly supervised training dataset for training a deep learning model. In the recognition phase, the HPO dictionary and the trained model are used for dictionary-based matching and a deep learning-based method, respectively, to recognize HPO concepts from biomedical texts. Finally, the results of the two taggers are combined with our rules. The details are described in the sections below.

### 2.1 Dictionary-based tagger

The dictionary-based method includes two processing steps: dictionary construction and text string-based matching. HPO is used as the dictionary resource to construct our phenotype dictionary. It is being increasingly adopted as a standard for phenotypic abnormalities by diverse groups, such as international rare disease organizations, registries, and clinical labs, as well as for biomedical resources and clinical software tools (Köhler *et al.*, 2019). Each term in HPO describes a clinical abnormality and is assigned to one of five subontologies (i.e., phenotypic abnormality, mode of inheritance, clinical modifier, clinical course, and frequency). Each term has a unique ID and a name label. Most terms have synonyms and textual definitions. For example, the concept named "limbal dermoid" has a unique ID (HP:0001140). Its synonyms include "benign eye tumor", "epibulbar dermoid", and "epibulbar dermoids".

We used the version of HPO released in 2019-11-08 to construct the phenotype dictionary. The HPO provides a standardized vocabulary of phenotypic abnormalities associated with 7,500+ human diseases and contains 14,000+ terms. All HPO concept names and their synonyms are extracted to build the dictionary. Note that the abbreviations of the terms are filtered, as they are often ambiguous (i.e., different concepts may have the same abbreviation). For example, "ASD" matches abbreviations for both "Atrial Septal Defect" and "Autism Spectrum Disorder". In addition, we used the Natural Language Toolkit (NLTK) (Bird *et al.*, 2009) to generate the lemmas of the concept names and synonyms to expand the dictionary, as it can obtain more general forms of the terms. At last, there are a total of 42,506 phenotype concept label names and synonyms with corresponding HPO IDs in our dictionary. Note that, when a concept text is matched to multiple different terms, it will be mapped to a new ID that concatenates all matched HPO IDs with semicolons. For example, the text "ear anomaly" is mapped to the ID (HP:0000356;HP:0000598).

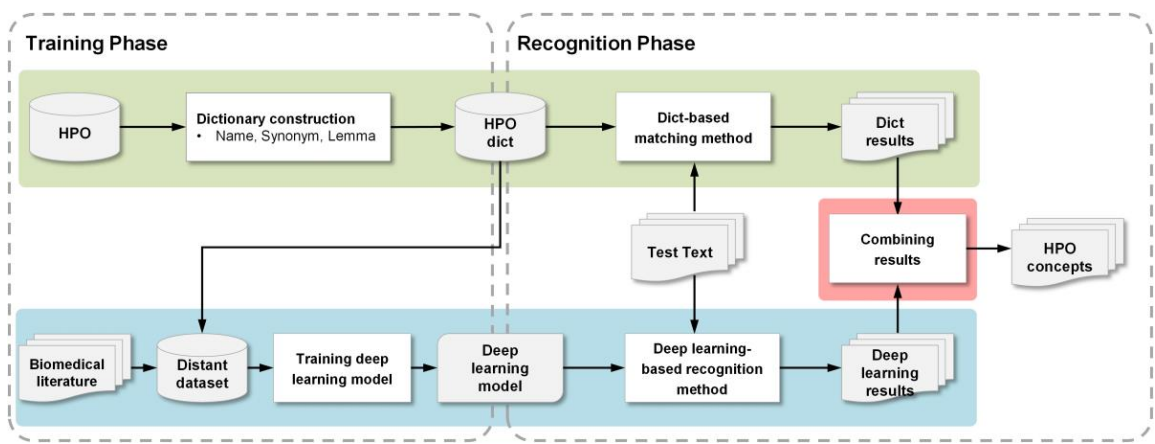

**Fig. 1.** Overview of PhenoTagger. The green, blue, and red parts denote the dictionary, deep learning-based tagger and combining result components, respectively.

To implement an efficient dictionary-based matching method, the Trie tree data structure (Fredkin, 1960) is adopted to store the HPO dictionary. The Trie tree is a multi-fork tree structure that can efficiently store dictionaries for rapid searching. Then, the prefix search is applied for exact matching using the case-insensitive mode. The dictionary-based method with exact matching can achieve a higher precision, as most of the concepts in the phenotype ontology are specific and not ambiguous. However, its main problem is that the concept name often has many spelling variant synonyms, leading to a lower recall rate. For example, the phenotype name "cupped ear" has variants such as "capuchin ear", "cup-shaped ear", "cup ear", and "cup-ear." Although some synonyms of the concept name are provided, it is impossible to cover all variations. The method cannot recognize the unseen concept synonyms in the dictionary. Therefore, we combined it with the following deep learning-based method to improve the recall.

## 2.2 Deep learning-based tagger

Different from previous pipelined concept recognition methods treating concept recognition as two separate tasks (i.e., NER results are first obtained and then NEN is used to link the recognized entity mentions to the corresponding concept labels), our deep learning-based method converts HPO concept recognition into a multi-class text classification task. Given a sequence of words (i.e., the text string of the concept name) as the input $x = \{x_1, x_2, ..., x_L\}$, the model classifies it into a class label of the concept ID $y \in [1, N]$, where $L$ is the length of the concept name and $N$ denotes the number of the predefined classes (i.e., the number of the unique HPO concept IDs). Then we used a deep learning-based text classification method to address the problem, which includes three processing steps: distantly supervised training dataset construction, deep learning model training, and concept recognition.

### 2.2.1 Distantly supervised training dataset construction

Compared with dictionary-based methods, machine learning-based methods can recognize more phenotype concept variants by automatic feature learning. However, most of them require large corpora of manually annotated data for model training. To tackle the HPO corpora scarcity issue, one approach is to use distant supervision to generate training data automatically. We leverage our HPO dictionary built by the method described in Section 2.1 and free PubMed Central (PMC) full text to automatically generate a distantly supervised training dataset to train our model.

Concretely, we paired each phenotype term (including label names and synonyms) and its concept ID in our dictionary as a training instance. The text of the phenotype term is tokenized using NLTK and converted to lowercase as the input of the model. Its corresponding HPO ID is used for the classification label as the output of the model. These instances in the HPO dictionary can be viewed as positives. In addition, we produced negatives from the biomedical literature for the texts that do not match any concept. Intuitively, the model can learn more useful information from the phenotype-related text description than unrelated text, and phenotype often is associated with disease and mutation. Therefore, we downloaded 150,052 PMC open access articles using the BioC API (Comeau *et al.*, 2019) with the query "disease and mutation". Then, we randomly sampled *n*-grams of words in the text after filtering out the positive terms as negatives, where *n* is a hyper-parameter (we select *n* from 1 to 10). The negatives are labeled with a new ID "HP: none". Although there is a chance that some HPO concept variants are included in the negative dataset, our method with negatives still achieves better performance than it does without negatives in our experiments.

### 2.2.2 BioBERT model

Recently, BERT (Bidirectional Encoder Representations from Transformers) (Devlin *et al.*, 2019), a contextualized word representation model that is pre-trained based on a masked language model using a deep bidirectional Transformer (Vaswani *et al.*, 2017), has shown promising results in a broad range of natural language processing (NLP) tasks and is widely used in the field of NLP (Peng *et al.*, 2019). We used the biomedical version of BERT (i.e., BioBERT) as our deep learning classifier. BioBERT (Lee *et al.*, 2020) is a biomedical language representation model pre-trained on large-scale biomedical corpora for biomedical text mining. Different from general domain texts, biomedical domain texts contain a considerable number of domain-specific terms and language structures. Therefore, BioBERT is initialized with weights from BERT and then is pre-trained on PubMed abstracts and PMC full-text articles for the biomedical domain. With minimal architectural modification, BioBERT can be applied to various downstream biomedical text mining tasks and significantly outperforms previous state-of-the-art models on the biomedical NER, relation extraction and question answering tasks.

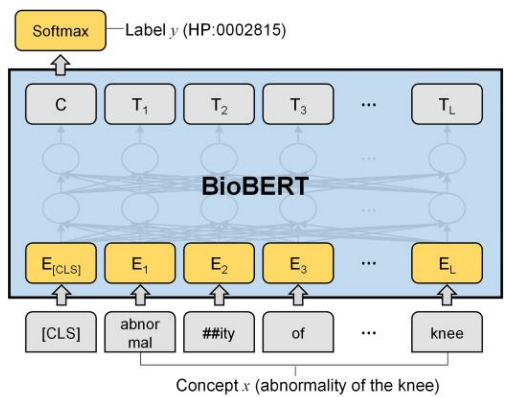

**Fig. 2.** Architecture of our BioBERT model

The architecture of our BioBERT model is illustrated in Fig. 2. Similar to the original version of BERT, WordPiece embeddings (Wu *et al.*, 2016) are used for BioBERT. The first token of every sequence is always a special classification token ([CLS]), for which a final hidden state vector $C$ is used as the aggregate concept sequence representation for the classification task. Then a fully connected layer with softmax activation is used as the output layer to predict the label probability scores $P = softmax(CW^T + b)$, where $W$ is the parameter of the output layer and $b$ is a bias. In the training phase, our BioBERT model is initialized with weights from BioBERT-Base v1.1 (https://github.com/dmis-lab/biobert) and then fine-tuned on HPO concept recognition task using our distantly supervised training dataset. All of the model parameters are optimized to maximize the log-probability of the correct label using the Adam optimizer (Kingma and Ba, 2015).

#### 2.2.3 Concept recognition

In the recognition phase, the input text is split into sentences, tokenized, and part-of-speech (POS) tagged, using NLTK, and converted to lowercase. All *n*-grams of words in the sentences are then generated as the concept candidates, where $n \in [2, 10]$. Note that, we do not use the deep learning-based tagger to predict the unigram concept candidates. There are two main reasons for this. First, unigram concepts have fewer variants so that the dictionary-based tagger can extract them well. Second, unigram candidates are easily recognized as false instances by the deep learning-based method due to the limited information provided by the text itself. For the maximum length of the *n*-grams, according to our statistics of the HPO terms, only less than 1% of the concepts in the HPO are longer than 10 words. Therefore, we selected 10 as the maximum length of the *n*-grams to make our method more efficient. Next, a filter based on the POS tags is used to filter the concept candidates. The filter removes the candidates that begin or end in punctuation, preposition conjunctions, subordinating conjunctions, coordinating conjunctions, and determiners. The trained BioBERT model is next used to classify each candidate into the concept ID. Finally, the candidates whose score (i.e., the label probability score $P$ predicted by the BioBERT model) is higher than the threshold $T$ are recognized as the HPO concepts, where $T$ is a hyper-parameter and is chosen based on the performance of the method on a development set.

### 2.3 Combining results

After the above annotation process, we can obtain the results from the dictionary and deep learning-based taggers. First, we assign each concept recognized by the dictionary-based tagger a score of 1. Then, we design the following four rules which consider overlapping concepts to combine the results of the two taggers:

(1) All non-overlapping concepts are retained.
(2) If the overlapping concepts have the same concept ID, the concept with the highest score is retained.
(3) If the overlapping concepts have the same start and end positions in text but are mapped into different concept IDs, the concept with the highest score is retained.
(4) If the overlapping concepts have different start and/or end positions in text and different concept IDs, all of the overlapping concepts are retained.

An example (the text of PMID 12592607) from the GSC+ test set for combining results is shown in Supplementary Materials A.1. After combining results, a post-processing method is used for abbreviation resolution, as the abbreviations were filtered in the previous step. Specifically, the abbreviation recognition algorithm (Schwartz and Hearst, 2003) is first used to identify abbreviations and their full names from the text, and then the abbreviations are tagged as the concept when their full names are already tagged.

## 3 Results

### 3.1 Experimental datasets and settings

In our experiments, the recently published HPO GSC+ (Lobo *et al.*, 2017) and a new dataset (named JAX, as it was created by the Jackson laboratory team) were used to evaluate the performance for the HPO concept recognition task. GSC+ consists of 228 manually annotated PubMed abstracts, with a total of 1,933 annotations that cover 497 unique HPO concepts. The JAX dataset provides only document-level annotations of HPO terms in 131 PMC full-text articles, with a total of 988 unique HPO concepts. In addition to HPO concept recognition, our method was also applied to disease concept recognition to test robustness and generalization. We used the disease ontology MEDIC to retrain PhenoTagger, and then evaluated it on the NCBI disease corpus (Doğan *et al.*, 2014). More detailed descriptions for each dataset are provided in Supplementary Material A.2. The precision (P), recall (R), and F1-score (F1), which are widely used in concept recognition tasks, are used to evaluate performance in our experiments. In particular, we used the document-level macro average metrics and mention-level micro average metrics to evaluate the prediction results. At the document level, only the set of concept ID labels within each document is considered, ignoring the exact concept positions in the text. Differently, at the mention level, every concept with their corresponding position in text is used to compute evaluation metrics. (More details are provided in Supplementary Materials A.3.)

For the parameter setting, we used BioBERT with the default parameter settings and tuned the other hyper-parameters of PhenoTagger on the development set by random search (Bergstra and Bengio, 2012). The main hyper-parameters are as follows: learning rate of 0.0001, batch size of 128, and threshold $T$ of 0.95. The number of training epochs is chosen by early stopping strategy (Prechelt, 1998) according to the performance on the development set (50 epochs at most). PhenoTagger was trained and tested on one NVIDIA Tesla V100 GPU. The training process for HPO takes ~16 hours to complete, and the testing on the GSC+ test set (206 PubMed abstracts) takes 152 seconds (see Supplementary Materials Table. S1 for details).

### 3.2 Performance comparison for phenotype concept

To demonstrate the effectiveness of PhenoTagger, we compare it with several state-of-the-art methods for phenotype concept recognition. (1) OBO (Taboada *et al.*, 2014): The OBO annotator was implemented specifically to annotate biomedical literature with HPO phenotypic abnormalities, which is based on the lexical and contextual matchings. (2) Doc2hpo (Liu *et al.*, 2019): We used Doc2hpo's ensemble engine, which combines the results generated from multiple string-based methods such as MetaMap server (Aronson, 2001). (3) MI (Shefchek *et al.*, 2020): The Monarch Initiative is an integrative data and analytic platform that connects phenotypes to genotypes across species. It allows a user to enter free text and perform an automated phenotype annotation on this text with terms from the Monarch knowledge graph. (4) NCR (Arbabi *et al.*, 2019): This is the ontology-guided neural concept recognizer based on a CNN. Note that not all methods are implemented with the entire HPO ontology, but they can identify the concepts in the phenotypic abnormality subontology. For example, OBO and NCR used only the phenotypic abnormality subontology to build their models. To ensure a fair comparison, we used only the concepts in the phenotypic abnormality subontology (the concepts in the subontology comprise 98% of all concepts in the ontology) to evaluate the results. We randomly selected 10% of GSC+ as the development set for hyper-parameter selection and early stopping. The remaining data are used as the GSC+ test set.

**Table 1.** Performance comparison with other existing methods on the GSC+ test set

| Method | Mention level | | | Document level | | |
|---|---|---|---|---|---|---|
| | P | R | F1 | P | R | F1 |
| OBO | **0.850** | 0.534 | 0.656 | **0.809** | 0.565 | 0.665 |
| Doc2hpo | 0.790 | 0.596 | 0.679 | 0.768 | 0.618 | 0.685 |
| MI | 0.799 | 0.617 | 0.696 | 0.757 | 0.605 | 0.673 |
| NCR | 0.789 | 0.589 | 0.674 | 0.741 | 0.602 | 0.664 |
| PhenoTagger | 0.789 | **0.722** | **0.754** | 0.774 | **0.740** | **0.757** |

Table 1 shows the evaluation results of the GSC+ test set including the overlapping concepts for the HPO concept recognition. The results show that PhenoTagger achieves the best F1 scores at both mention and document levels. Among other methods, OBO, Doc2hpo and MI are based on dictionary matching, and OBO achieves the highest precision. However, these methods suffer from low recall rates since they have difficulty in effectively recognizing any unseen concept synonyms. Although NCR can recognize more new concept synonyms using the CNN model, it cannot recognize the overlapping concepts that lead to the low recall. In contrast, PhenoTagger is a hybrid method that leverages the advantages of the dictionary and deep learning-based methods. It can not only exactly match the concepts in the dictionary, but it can also identify more unseen concept synonyms using the deep learning model. Additionally, it can extract the overlapping concepts due to our combined rules. The results show that recall is improved significantly without loss of precision. PhenoTagger achieves improvements of 0.105 and 0.122 in recall as compared to other methods at the mention level and document level on the GSC+ test set, respectively.

Because OBO and NCR do not recognize the overlapping concepts, the comparison results for non-overlapping concepts on the GSC+ test set are also shown in Table 2. Excluding OBO and NCR, other methods filter overlapping concepts by choosing the longest ones as the result of the non-overlapping concept. Compared with the dictionary-based methods (i.e., row 1-3), NCR achieves the higher recall. It uses a CNN to encode query phrases into vector representations and computes their similarity to embeddings learned for ontology concepts. Thus, it can recognize more concept synonyms by automatic feature learning. However, there are obvious drops in precision at both levels so that NCR does not achieve better performance than Doc2hpo at the document level. The main reason is that using only a deep learning-based method cannot guarantee the precision of the predictions due to the limitations of the training dataset. Some false positives with high similarity scores are introduced, especially for the concept of the unigram. Different from NCR, PhenoTagger uses the BioBERT model as our deep learning-based tagger, and the model can achieve better performance than does the CNN model (We also tested the performance of the PhenoTagger with different deep learning models, including CNN, recurrent neural network (RNN), Transformer, and BERT. The results are provided in Supplementary Materials A.4). Moreover, PhenoTagger combines the dictionary-based results and the deep learning-based prediction results with higher scores for improved performance. Therefore, PhenoTagger achieves the best performance for non-overlapping concepts.

**Table 2.** Performance comparison with other existing methods for non-overlapping concepts on the GSC+ test set

| Method | Mention level | | | Document level | | |
|---|---|---|---|---|---|---|
| | P | R | F1 | P | R | F1 |
| OBO | 0.777 | 0.579 | 0.664 | 0.746 | 0.590 | 0.659 |
| Doc2hpo | 0.749 | 0.638 | 0.689 | 0.742 | 0.632 | 0.683 |
| MI | 0.777 | 0.587 | 0.669 | 0.734 | 0.577 | 0.646 |
| NCR | 0.721 | **0.695** | 0.708 | 0.691 | 0.671 | 0.681 |
| PhenoTagger | **0.784** | **0.695** | **0.737** | **0.779** | **0.713** | **0.745** |

Further, we test PhenoTagger on a new JAX dataset. In the dataset, only HPO concepts for special patients are annotated without the locations. Therefore, we focus only on the recall at the document level, as exact precision cannot be obtained. The performance comparison on the JAX dataset is shown in Fig.3. Consistent results are observed when conducting testing on the independent benchmarking dataset. Compared with other methods, PhenoTagger can identify more phenotype concepts and achieve a higher recall. In summary, the results on the GSC+ and JAX datasets demonstrate the effectiveness of PhenoTagger for phenotype concept recognition.

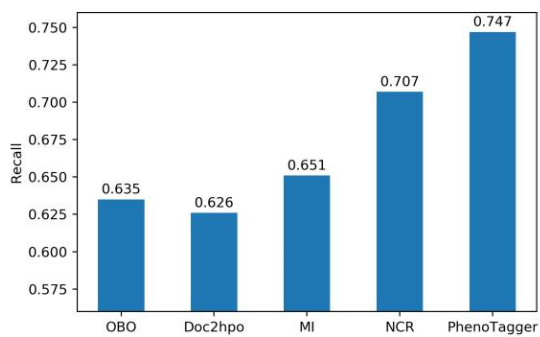

**Fig. 3.** Document-level Macro Average recall on the JAX dataset

### 3.3 Performance comparison for disease concept

In this section, to evaluate the effectiveness of our method on the different ontologies, PhenoTagger is applied to a disease ontology for disease concept recognition. Specifically, we retrained our model for disease concept recognition using the disease ontology MEDIC and evaluated the performance on the NCBI disease test set. The performance comparison with other state-of-the-art methods is shown in Table 3.

**Table 3.** Performance comparison with other existing methods on the NCBI Disease test set

|  | Mention level | | | Document level | | |
|---|---|---|---|---|---|---|
| Method | P | R | F1 | P | R | F1 |
| Dict | 0.776 | 0.588 | 0.669 | 0.827 | 0.677 | 0.745 |
| DNorm | 0.788 | 0.761 | 0.774 | 0.823 | 0.817 | 0.820 |
| TaggerOne | **0.835** | **0.828** | **0.831** | 0.847 | **0.836** | **0.841** |
| PhenoTagger | 0.815 | 0.695 | 0.750 | **0.852** | 0.789 | 0.819 |

Among other methods, Dict is our dictionary-based method described in Section 2.1, which uses the disease ontology instead of the original HPO. Similar results with phenotype concept recognition are observed, and PhenoTagger can improve recall significantly without loss of precision and achieve better performance than does the dictionary-based method. DNorm (Leaman *et al.*, 2013) and TaggerOne (Leaman and Lu, 2016) are based on supervised machine learning methods and trained by the NCBI disease training dataset. DNorm is a pipeline method, which first locates the disease mentions using the BANNER recognizer (Leaman and Gonzalez, 2008), it then normalizes each mention to a MEDIC concept using pairwise learning-to-rank. Different from DNorm, TaggerOne is a joint learning method for biomedical entity recognition and normalization with semi-Markov models, which can overcome cascading errors and fully exploit dependencies of the entity recognition and normalization. For DNorm and TaggerOne, we first obtained the prediction results on the NCBI Disease test set using the official tools, and then computed our evaluation metrics. At the document level, PhenoTagger without manually labeled training data achieves competitive performance compared with the state-of-the-art supervised methods. The results suggest that our method is highly robust and generalizable. At the mention-level, PhenoTagger obtains lower recall. The main reason is that it does not identify some abbreviations of concepts, whereas the abbreviations appear many times in the NCBI disease corpus. As described in Section 2.3, PhenoTagger tags only the abbreviations as the concept when their full names already are tagged, leading to error propagation from the abbreviation recognition algorithm.

### 3.4 Analysis

#### 3.4.1 Effect of the different numbers of negatives

As described in Section 2.2.1, we used 42,506 concept label names and synonyms in the HPO dictionary as positives and randomly sampled some *n*-grams from the PMC full text as negatives to build our distantly supervised training dataset. To explore the effect of negatives on performance, we conducted the experiment with different numbers of negatives (i.e., 0, 5,000, 10,000, 50,000, 100,000 and 200,000). The performance of PhenoTagger on the HPO GSC+ test set is shown in Fig. 4.

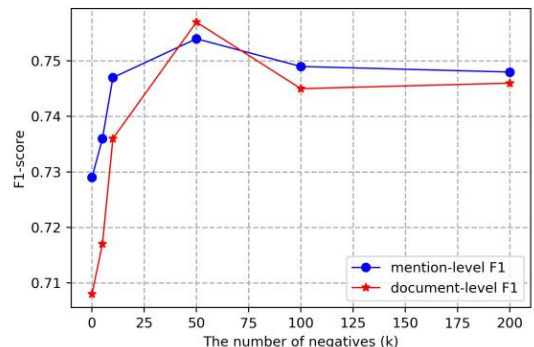

**Fig. 4.** Effect of the different numbers of negatives on the GSC+ test set

With the number of negatives increasing, the performance of PhenoTagger is gradually improved. When the number of negatives is 50,000, the best performance with the F-scores of 0.754 and 0.757 (improvements of 0.025 and 0.049 over the model without negatives) are achieved on the GSC+ test set at the mention and document levels, respectively. However, a slight performance drop is observed when more negatives are used to train the model. Too many negatives may lead to an unbalanced class distribution, so that the positives are more difficult to learn. Further, more negatives do not provide additional useful information for model training. This suggests that PhenoTagger trained on a balanced version of distantly supervised training dataset can achieve better performance. In addition to negatives generated from biomedical literature, we also used random *n*-grams from unrelated text (i.e., Wikipedia) as negatives but observed worse performance (F-scores of 0.733 and 0.718 at the mention and document levels, respectively). This suggests the negatives generated from related domain text are more useful than that from unrelated text.

#### 3.4.2 Effect of different tagger components

To further analyze the effectiveness of dictionary and deep learning-based tagger components for our hybrid method, the performance of each individual tagger was tested. Table 4 presents the evaluation results of the different taggers on the HPO GSC+ test set. Here, Dict denotes the dictionary-based tagger described in Section 2.1. DL denotes the deep learning-based tagger described in Section 2.2, but the difference is that unigrams of words are also generated as the concept candidates in the recognition phase. Hybrid denotes our complete hybrid method described in Section 2. Note that, the individual deep learning-based model of DL was retrained independently.

**Table 4.** Performance of different components on the GSC+ test set

|  | Mention level | | | Document level | | |
|---|---|---|---|---|---|---|
| Method | P | R | F1 | P | R | F1 |
| Dict | **0.834** | 0.582 | 0.686 | 0.756 | 0.565 | 0.647 |
| DL | 0.784 | 0.666 | 0.720 | 0.750 | 0.695 | 0.721 |
| Hybrid | 0.789 | **0.722** | **0.754** | **0.774** | **0.740** | **0.757** |

As seen in Table 4, our dictionary-based tagger obtains high precision but suffers from lower recall, which is similar to those of other existing dictionary-based methods. The deep learning-based tagger achieves better F1-scores and a significant improvement in recall, but a performance drop in precision is observed. In practice, the performance of machine learning models often depends on the labeled training corpus. Due to the limitations of the manually annotated dataset for HPO concepts, we used the distantly supervised training dataset built with the HPO to train the model. Based on our statistics for the HPO terms (as shown in Supplementary Materials Fig. S2), there are over 4,000 concepts without synonyms. Compared to the concepts with enough synonyms, the deep learning model has difficulty with correctly identifying unseen variants of the concept without synonyms, as there are not enough training instances for feature learning. Moreover, unigram candidates are easily recognized as false instances by the deep learning-based tagger due to the limited information provided by the text itself. When the two methods are combined, the hybrid method achieves better performance. In summary, the deep learning method can alleviate the low recall problem of the dictionary-based method, and the dictionary-based method can help the deep learning-based method to recognize the concepts without sufficient training information. Therefore, the two methods are complementary, and combining them can help to improve performance.

### 3.4.3 Error analysis

Although PhenoTagger exhibits promising results for phenotype concept recognition, there are remaining questions to address. To understand the causes of errors made by PhenoTagger, we manually analyzed a random sample of both HPO corpora for errors and describe the trends observed. We found PhenoTagger can correctly identify some entity mentions from text, but it assigns the wrong HPO IDs. This error is mainly caused by two reasons. First, PhenoTagger cannot disambiguate the different concepts with the same text name. For example, the concept "abnormality of the outer ear" (HP:0000356) and the concept "abnormality of the ear" (HP:0000598) have the same synonym "ear anomaly". According to our statistics, only 0.05% of the concept synonyms in the HPO have multiple different term IDs. Second, some concepts are so similar that they are easily confused (e.g., "generalized hypopigmentation" (HP:0007513) vs. "hypopigmentation of the skin" (HP:0001010), "hypotonia in infancy" (HP:0008947) vs. "neonatal hypotonia" (HP:0001319)). Currently, PhenoTagger only uses the text of a concept name as input, while the text of some concept names does not provide sufficient information to allow the model to reliably differentiate between them. Further, we also observed that PhenoTagger often misses some phenotype concepts that are lexically dissimilar with the concept terms in the HPO. For example, the text "jaw cyst" in PMID 6882181 fails to be recognized as the phenotype concept (HP:0010603), because it is lexically dissimilar with any of the terms representing the concept of HP:0010603, such as "odontogenic keratocysts of the jaw" and "keratocystic odontogenic tumor". To alleviate these problems, only the information of concept names and synonyms is insufficient, and additional information would be needed. HPO also provides the taxonomic hierarchy structure information (i.e., "is_a" relations). The prior knowledge provides additional semantic information for HPO concepts. For example, "Abnormality of the genitourinary system" (HP:0000119) is the ancestor of "urinary tract atresia" (HP:0000809); they are lexically dissimilar but semantically similar. More recently, HPO2Vec+ has been proposed to leverage heterogeneous knowledge resources to enrich the HPO embeddings (Shen *et al.*, 2019). Incorporating such taxonomic structure information and embeddings into the model holds great potentials by sharing knowledge between related concepts and providing prior semantic information on the similarity between concepts. Moreover, the concept definitions provided by HPO and the contexts surrounding the concepts in the text may also provide useful information. How to apply such additional information for further model enhancement will be explored in our future work.

## 4 Conclusion

In this paper, we propose PhenoTagger, a novel hybrid phenotype concept recognition approach that combines dictionary and deep learning-based taggers to utilize their complementary characteristics. We tested PhenoTagger on the GSC+ corpus and a new JAX corpus for phenotype concept recognition. The results show that PhenoTagger compares favorably to the state-of-the-art methods on the two corpora. The PhenoTagger also was retrained using the disease ontology MEDIC, and tested on the NCBI disease corpus for disease concept recognition. PhenoTagger achieves competitive performance as compared with state-of-the-art fully supervised methods. Our experiments demonstrate the effectiveness of our ontology-driven PhenoTagger without requiring manually labeled training data.

PhenoTagger exhibits promising results for phenotype concept recognition and can be easily adapted to other biomedical concepts. Nevertheless, it has several limitations and provides opportunities for further research. Unlike other fully supervised NER methods that can identify a novel entity according to the context of the text, PhenoTagger can identify only the concepts in the ontology and cannot discover the novel concepts. Further, PhenoTagger cannot disambiguate different concepts with the same name, as it uses only the text of concept names as input. We will explore the use of additional information, such as HPO taxonomic information, HPO concept definitions and the context of concepts in the text, to address these problems in our future work. In addition, the processing speed of the PhenoTagger is relatively slow compared to other existing methods due to the complexity of the BioBERT model. Using knowledge distillation technology to speed up the processing of the model is another direction for future work.

## Acknowledgements

This research is supported by the Intramural Research Programs of the National Institutes of Health, National Library of Medicine.
Thanks to Chih-Hsuan Wei for his help with Web APIs.
*Conflict of Interest:* none declared.

# PhenoTagger: A Hybrid Method for Phenotype Concept Recognition using Human Phenotype Ontology
# (Supplementary Materials)

## A.1 Example for combining results

Our combining rules:

(1) All non-overlapping concepts are retained.
(2) If the overlapping concepts have the same concept ID, the concept with the highest score is retained.
(3) If the overlapping concepts have the same start and end positions in text but are mapped into different concept IDs, the concept with the highest score is retained.
(4) If the overlapping concepts have different start and/or end positions in text and different concept IDs, all of the overlapping concepts are retained.

An example (the text of PMID 12592607) from the GSC+ test set for combining results is shown in Fig. S1. In this case, the non-overlapping concept of (4, 25, distal arthrogryposes, HP:0005684, 0.999) is retained by Rule 1. For the overlapping concepts in position indexes from 74 to 106, the concepts of (74, 106, multiple congenital contractures, HP:0002804, 1.000) and (83, 106, congenital contractures, HP:0002803, 1.000) with the highest scores are retained by Rule 2, and (94, 106, contractures, HP:0001371, 1.000) is retained by Rule 4 due to the different position and the different concept ID. For the overlapping concepts in position indexes from 428 to 434, (428, 434, twitch, HP:0010546, 1.000), the one with the higher score is retained by Rule 2. Similarly, the concept of (1048, 1054, twitch, HP:0010546, 1.000) is retained. Finally, for the overlapping concepts in position indexes from 1149 to 1180, (1149, 1180, multiple-congenital-contracture, HP:0002804, 0.9999), (1158, 1180, congenital-contracture, HP:0002803, 0.99999) and (1169, 1180, contracture, HP:0001371, 1.000) are retained by Rule 4.

The results of dictionary-based and deep learning-based taggers

| Start | End | Text | HPO_ID | Score | Tagger |
|---|---|---|---|---|---|
| 4 | 25 | distal arthrogryposes | HP:0005684 | 0.999 | DL |
| 74 | 106 | multiple congenital contractures | HP:0002804 | 1.000 | Dict |
| 74 | 106 | multiple congenital contractures | HP:0002804 | 0.999 | DL |
| 83 | 106 | congenital contractures | HP:0002803 | 1.000 | Dict |
| 83 | 106 | congenital contractures | HP:0002803 | 0.999 | DL |
| 94 | 106 | contracture | HP:0001371 | 1.000 | Dict |
| 428 | 434 | twitch | HP:0010546 | 1.000 | Dict |
| 428 | 444 | twitch myofibers | HP:0010546 | 0.956 | DL |
| 1048 | 1054 | twitch | HP:0010546 | 1.000 | Dict |
| 1048 | 1064 | twitch myofibers | HP:0010546 | 0.956 | DL |
| 1149 | 1180 | multiple-congenital-contracture | HP:0002804 | 0.999 | DL |
| 1158 | 1180 | congenital-contracture | HP:0002803 | 0.999 | DL |
| 1169 | 1180 | contracture | HP:0001371 | 1.000 | Dict |

The combined results

| Start | End | Text | HPO_ID | Score |
|---|---|---|---|---|
| 4 | 25 | distal arthrogryposes | HP:0005684 | 0.999 |
| 74 | 106 | multiple congenital contractures | HP:0002804 | 1.000 |
| 83 | 106 | congenital contractures | HP:0002803 | 1.000 |
| 94 | 106 | contractures | HP:0001371 | 1.000 |
| 428 | 434 | twitch | HP:0010546 | 1.000 |
| 1048 | 1054 | twitch | HP:0010546 | 1.000 |
| 1149 | 1180 | multiple-congenital-contracture | HP:0002804 | 0.999 |
| 1158 | 1180 | congenital-contracture | HP:0002803 | 0.999 |
| 1169 | 1180 | contracture | HP:0001371 | 1.000 |

**Fig. S1.** Example for combining results of dictionary-based and deep learning-based taggers (PMID 12592607). Dict and DL denotes the dictionary and deep learning-based taggers, respectively.

## A.2 Corpora

In our experiments, the recently published HPO GSC+ (Lobo et al., 2017) and a new dataset (JAX) were used to evaluate the performance for the HPO concept recognition task. (1) GSC+: Groza et al. (2015) provided a unique gold standard corpus (GSC) for HPO. GSC+ is an extended version of the GSC that added 881 entities and modified 4 entities (Lobo et al., 2017). The dataset consists of 228 manually annotated PubMed abstracts, with a total of 1,933 annotations that cover 497 unique HPO concepts. GSC+ provides mention-level annotations of HPO terms. Note that the overlapping concepts also are annotated in the dataset (about 26% of the concepts are overlapping concepts). We randomly selected 10% of the dataset as the development set for hyper-parameter selection and early stopping. The remaining data are used as the test set. (2) JAX: This is a new dataset and created by the Jackson laboratory team. The dataset provides only document-level annotations of HPO terms without the positions of the mentions; it consists of 131 PMC full-text articles, covering 988 unique HPO concepts. In the dataset, only HPO concepts for special patients are annotated; other general HPO concepts are not annotated. In addition to HPO concept recognition, our method was also applied to disease concept recognition to test robustness and generalization. We used the disease ontology MEDIC to retrain PhenoTagger, and then evaluated it on the NCBI disease corpus (Doğan et al., 2014). The NCBI disease corpus consists of 793 PubMed abstracts and is split into three subsets (i.e., training, development, and test sets). The development set was used for hyper-parameter selection and early stopping, and the final evaluation was performed using the test set.

### A.3 Evaluation metrics

The precision (P), recall (R), and F1-score (F1), which are widely used in concept recognition tasks, are used to evaluate performance in our experiments. In particular, we used two versions of the metrics (i.e., the document-level macro average metrics and mention-level micro average metrics) to evaluate the prediction results. First, the document-level macro average metrics are used to evaluate the prediction results, which are calculated as follows:

$$P_{Macro} = \frac{1}{|D|} \sum_D \frac{TP_d}{TP_d + FP_d}$$

$$R_{Macro} = \frac{1}{|D|} \sum_D \frac{TP_d}{TP_d + FN_d}$$

$$F1_{Macro} = \frac{2 * P_{Macro} * R_{Macro}}{P_{Macro} + R_{Macro}}$$

where $TP_d$, $FP_d$, and $FN_d$ denote the number of true positives, false positives, and false negatives in document *d*, respectively; *D* denotes the total number of documents. Note that, when the number of prediction concepts and gold concepts for the document *d* are zero, we assign a macro recall and macro precision of 1.0, respectively. At the document level, only the set of concept ID labels within each document is considered, ignoring the exact concept positions in the text. Second, the mention-level micro average metrics are used for the evaluation. Differently, at the mention level, every concept with their corresponding position in text is used to compute evaluation metrics . This requires not only the correct concept ID label but also span matching (here, we use the relaxed span matching measure that only requires the spans to share at least one token). These metrics are calculated as follows:

$$P_{Micro} = \frac{\sum_D TP_d}{\sum_D TP_d + \sum_D FP_d}$$

$$R_{Micro} = \frac{\sum_D TP_d}{\sum_D TP_d + \sum_D FN_d}$$

$$F1_{Micro} = \frac{2 * P_{Micro} * R_{Micro}}{P_{Micro} + R_{Micro}}$$

## A.4 Performance of different deep learning models

In this section, we tested the performance of the PhenoTagger with different deep learning models (including CNN, recurrent neural network (RNN), Transformer, and BERT). For these models (excluding BERT and BioBERT), we used the concatenation of the 200-dimensional pre-trained BioWordVec (Zhang et al., 2019) and the character embeddings learned from a CNN layer followed by a max-pooling layer as the inputs. Then, CNN (Kim, 2014), bidirectional Long Short-Term Memory (Hochreiter and Schmidhuber, 1997) (BiLSTM), and self-attention-based Transformer (Vaswani et al., 2017) layers are used to extract the text features in the CNN, RNN, and Transformer models, respectively. Afterward, a max-pooling layer is used to extract global features. Finally, a fully connected layer with a softmax function is used as the output layer to classify text. For the BERT model, we used the official BERT-Base model (https://github.com/google-research/bert), which has the same neural network architecture as does BioBERT. Table S1 shows the results of PhenoTagger with the different deep learning models on the GSC+ test set. Moreover, the training/test time on one NVIDIA Tesla V100 GPU is provided.

**Table S1.** Performance of PhenoTagger with different deep learning models on the GSC+ test set

| Method | Training/Test time | Mention level | | | Document level | | |
|---|---|---|---|---|---|---|---|
| | | P | R | F1 | P | R | F1 |
| PhenoTagger (CNN) | $2h56m/106s$ | 0.772 | 0.706 | 0.738 | 0.735 | 0.706 | 0.720 |
| PhenoTagger (BiLSTM) | $4h4m/101s$ | 0.785 | 0.666 | 0.721 | 0.748 | 0.663 | 0.703 |
| PhenoTagger (Transformer) | $2h46m/96s$ | 0.770 | 0.696 | 0.731 | 0.741 | 0.707 | 0.724 |
| PhenoTagger (BERT) | $15h42m/152s$ | 0.791 | 0.700 | 0.743 | 0.757 | 0.705 | 0.730 |
| PhenoTagger (BioBERT) | $15h42m/152s$ | 0.789 | 0.722 | 0.754 | 0.774 | 0.740 | 0.757 |

Here, $h$, $m$, $s$ denotes hour, minute and second, respectively. Since BERT and BioBERT modes have the identical model architecture, their training/test time is the same.

The results show that, among the models without the pre-trained process, the CNN and Transformer models achieve similar performance and both outperform the BiLSTM model. The main reason is that our concept classification mainly relies on the string text of the concept itself rather than the long-distance dependent information. Compared with BiLSTM, CNN and Transformer can more effectively capture the local features of the concept text. For the pre-trained models, although the model architectures of BERT and BioBERT are similar to that of Transformer, they achieve better performances than does Transformer without the pre-trained process. The pre-trained models are more effective on the small training dataset. Different from the original BERT model which was pre-trained on general domain corpora (English Wikipedia and BooksCorpus), BioBERT is a pre-trained biomedical language representation model on PubMed abstracts and PMC full-text articles for biomedical text mining and achieves the best performance. The main reason is that biomedical domain texts contain a considerable number of domain-specific terms and language structures that are different from general domain text. The in-domain BioBERT can learn a contextualized word representation that contains more useful biomedical background and linguistic information by pre-training on the large biomedical literature. Although PhenoTagger with BioBERT needs substantially more time for training and testing due to the complexity of the BERT model, it achieves significant better performance in terms of accuracy.

**Fig. S2.** Distribution of the number of synonyms per positive concept label in HPO

*L.Luo et al.*

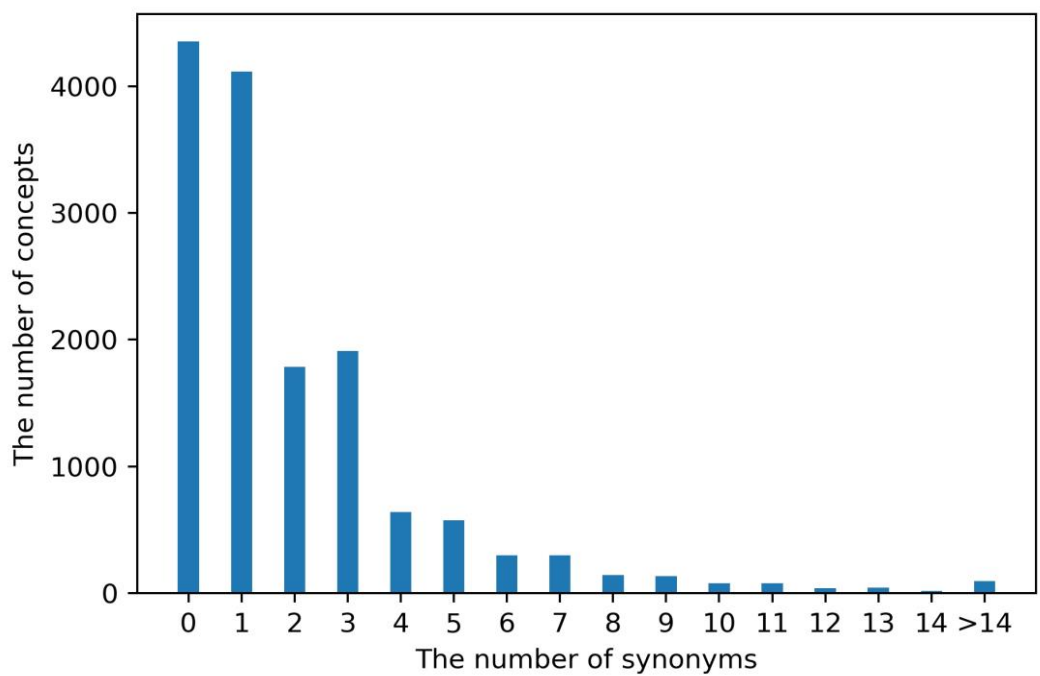

## Acknowledgements

This research is supported by the Intramural Research Programs of the National Institutes of Health, National Library of Medicine.

Conflict of Interest: none declared.